\PassOptionsToPackage{compatibility=false}{caption}

\documentclass{article}

\usepackage{microtype}
\usepackage{graphicx}
\usepackage{booktabs}
\usepackage{amsmath,amssymb}
\usepackage{multirow}

\usepackage{hyperref}

\usepackage{caption}
\usepackage{subcaption}

\usepackage[accepted]{icml2025}

\usepackage{amsmath}
\usepackage{amssymb}
\usepackage{mathtools}
\usepackage{amsthm}

\usepackage[capitalize,noabbrev]{cleveref}

\theoremstyle{plain}

\theoremstyle{definition}

\theoremstyle{remark}

\usepackage[textsize=tiny]{todonotes}

\icmltitlerunning{Direct Robot Configuration Space Construction using Convolutional
Encoder-Decoders}

\begin{document}

\twocolumn[
\icmltitle{Direct Robot Configuration Space Construction using Convolutional
Encoder-Decoders}



\icmlsetsymbol{equal}{*}
\begin{icmlauthorlist}
\icmlauthor{Christopher Benka}{yyy}
\icmlauthor{ Judah Goldfeder}{yyy}
\icmlauthor{Carl Gross}{yyy}
\icmlauthor{Riya Gupta}{yyy}
\icmlauthor{Hod Lipson}{comp}
\end{icmlauthorlist}

\icmlaffiliation{yyy}{Columbia University Department of Computer Science}
\icmlaffiliation{comp}{Columbia University Data Science Institute}

\icmlcorrespondingauthor{Firstname1 Lastname1}{first1.last1@xxx.edu}
\icmlcorrespondingauthor{Firstname2 Lastname2}{first2.last2@www.uk}

\icmlkeywords{Machine Learning, ICML}

\vskip 0.3in
]




\begin{abstract}
Intelligent robots must be able to perform safe and efficient motion planning in their environments. Central to modern motion planning is the configuration space. Configuration spaces define the set of configurations of a robot that result in collisions with obstacles in the workspace, $\text{C}_{\text{clsn}}$, and the set of configurations that do not, $\text{C}_{\text{free}}$. Modern approaches to motion planning first compute the configuration space and then perform motion planning using the calculated configuration space. Real-time motion planning requires accurate and efficient construction of configuration spaces. 
We are the first to apply a convolutional encoder-decoder framework for calculating highly accurate approximations to configuration spaces, essentially learning how the robot and physical world interact. Our model achieves an average 97.5\% F1-score for predicting $\text{C}_{\text{free}}$ and $\text{C}_{\text{clsn}}$ for 2-D robotic workspaces with a dual-arm robot. Our method limits undetected collisions to less than 2.5\% on robotic workspaces that involve translation, rotation, and removal of obstacles. Our model learns highly transferable features between  robotic workspaces, requiring little to no fine-tuning to adapt to new transformations of obstacles in the workspace.

\end{abstract}

\section{Introduction}
At the heart of robotic motion planning is the configuration space, hereafter referred to as the C-space. The C-space is comprised of all configuration vectors, $\vec{q}$, of a robot. Each dimension of the vector, $\vec{q}_i$, specifies the configuration of the ith degree of freedom (DOF) of the robotic system. A robot with n-DOFs will have a C-space of n dimensions. Industrial robotic systems typically have five or six DOF to perform complicated tasks like welding and palletizing. 
\begin{figure}[b]
    \centering
    \includegraphics[scale=.22]{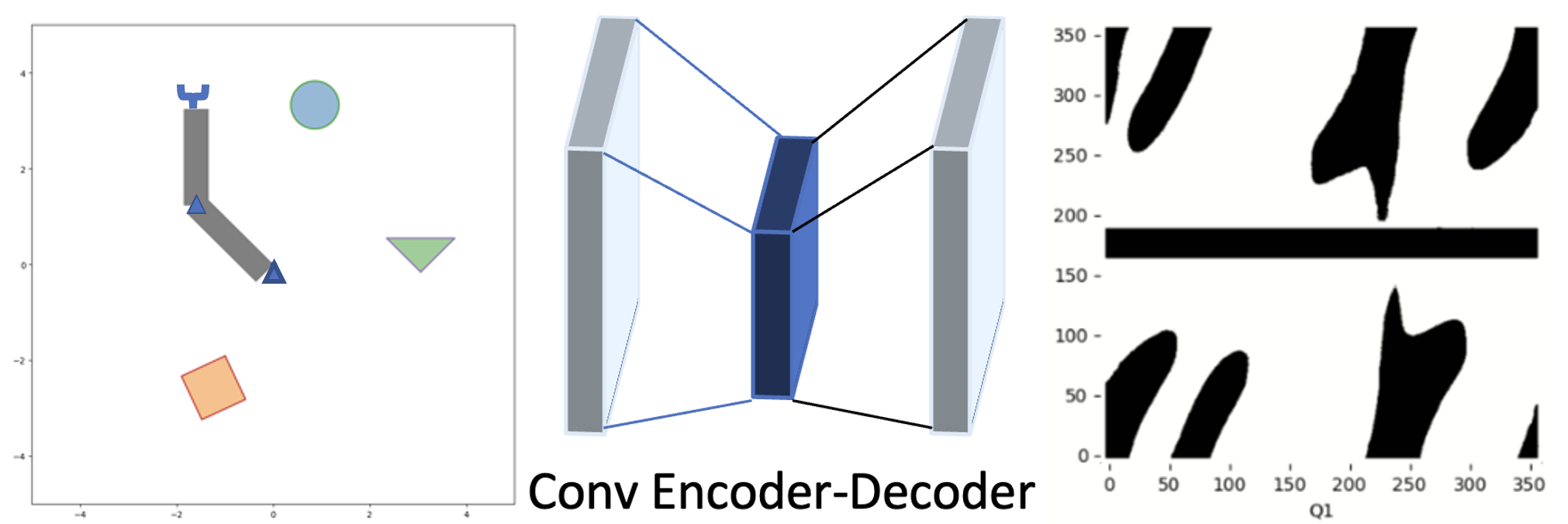}
    \caption{Technique visualized. Our convolutional encoder-decoder constructs highly approximate C-spaces for dynamic, 2-D workspaces involving translation, rotation, and removal of obstacles for a dual-arm robot. Our method computes the approximate C-space in 283 milliseconds on average, or $1.08 \text{ microseconds } (\mu s)$ per configuration.}
    \label{robotic-workspace-cspace-pairs}
\end{figure}
The C-space can be decomposed into two disjoint sets, $\text{C}_{\text{free}}$ and $\text{C}_{\text{clsn}}$ whose union completely defines the C-space. $\text{C}_{\text{free}}$ and $\text{C}_{\text{clsn}}$ denote the sets of configurations for which the robot does and does not collide with obstacles in the workspace, respectively. Since the two sets are disjoint, to compute the C-space, one needs only compute the boundary between $\text{C}_{\text{clsn}}$ or $\text{C}_{\text{free}}$ \cite{b4}. Once computed, using the C-space for path planning reduces the task from motion planning of a dimensional object to motion planning of a point \cite{b4}. The motion planning system uses the C-space to define a trajectory of configurations $q_t$ $\in \text{C}_{\text{free}}$ that takes the robot from its initial configuration $\text{q}_{\text{initial}}$  to end configuration $\text{q}_{\text{goal}}$. This trajectory is often referred to as a free-path \cite{b4}.

Efficiently computing accurate approximations of C-spaces remains an open problem. For robots with a large number of DOF, the boundary between $\text{C}_{\text{free}}$ and $\text{C}_{\text{clsn}}$ is highly complex. Sampling-based motion planning (SBMP) methods like probabilistic roadmaps (PRM) \cite{b1} and rapidly-exploring random trees (RRT) \cite{b2} are used to efficiently compute an approximation to the boundary. Recent works have focused on improving the efficiency of SBMP methods by replacing traditional, exact collision-checking procedures with approximate collision-checking through machine learning methods \cite{b3}. The machine learning techniques learn a classifier f($\vec{q}$) which maps a configuration $\vec{q}$ to either the set $\text{C}_{\text{free}}$ or $\text{C}_{\text{clsn}}$ with high accuracy.

For each robotic workspace, the classifier is trained on 10k to 100k sample $(\vec{q},y)$ pairs, where $y \in \{{1,-1}\}$ denoting collision and no-collision, respectively. Anytime a obstacle undergoes translation or rotation, or the number of obstacles in the workspace changes, the classifier must be retrained on another 10k to 100k samples of the new workspace \cite{b3}. 

In this work, we bypass SBMP methods by proposing direct C-space construction through convolutional encoder-decoders. We learn a function F that directly maps an image of a robotic workspace to an image of an accurate approximation of the C-space. Since the model is trained on robotic workspace, C-space pairs, the model can construct C-spaces for dynamic workspaces that include non-stationary obstacles and add or remove obstacles without the need to be retrained. Any configuration $q_t$ in the free-path $[\text{q}_{\text{initial}}, \dots, \text{q}_{\text{goal}}]$, computed using approximate C-space, can be checked using an exact collision detection procedure \cite{b3}. For any configurations $q_t$ in the computed free-path that result in a collision, a local repairing operation can be performed to update the path \cite{b3}. The configuration $q_t$ in the computed C-space can be updated to be included in $\text{C}_{\text{clsn}}$, avoiding any repeated local repairing operations on configuration $q_t$.

\begin{figure*}[]
\vspace*{2mm}
    \centering
        \includegraphics[scale=.16]{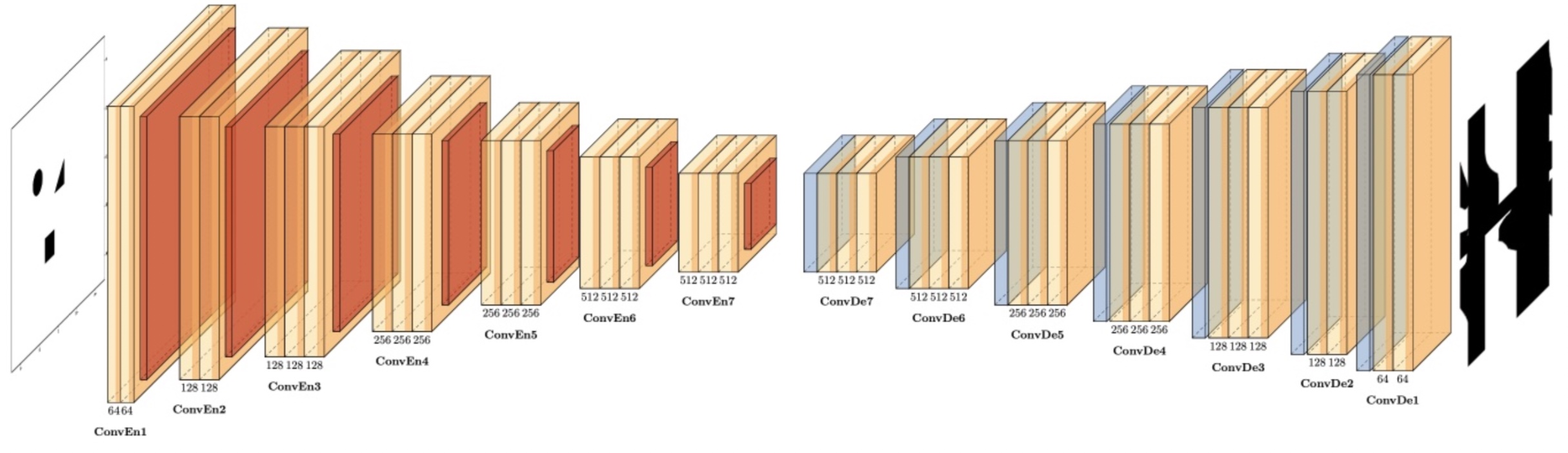}
    \label{model-arch}
    \caption{Illustration of model architecture. Modified from the original SegNet architecture, our convolutional encoder-decoder utilizes 7 encoders and 7 decoders for a total of 38 convolutional layers.}
\end{figure*}

Overall, our contributions are summarized as follows: 
\begin{itemize}
    \item We are the first to construct C-spaces directly from robotic workspaces using a convolutional encoder-decoder. The model achieves a $97.5\%$ F1-score for the prediction of configurations that belong to $\text{C}_{\text{free}}$ and $\text{C}_{\text{clsn}}$ and limits undetected collisions to less than $2.5\%$.
    \item Our model learns highly transferable features between robotic workspaces involving new transformations on obstacles. After training on translation of obstacles, the model adapts to the removal and rotation of obstacles with little to no fine-tuning.
\end{itemize}

\section{RELATED WORK}
Our work is related to those who have applied machine learning to build an understanding of a robotic workspace's C-space. Recent machine learning methods focus on improving collision detection in SBMP techniques. Garcia et al. developed a feed-forward network for detecting collisions between objects that have similar shape \cite{b9}. Pan and Manocha developed a support vector machine (SVM) classifier for learning an approximate boundary between $C_\text{clsn}$ and $C_\text{free}$ \cite{b17}. They show that as the number of samples increases, the predicted $C_\text{clsn}$ converges to the exact $C_\text{clsn}$ space \cite{b17}. Other techniques for learning a classifier that approximates the boundary have been proposed including, k-nearest neighbors (K-NN) \cite{b18,pan2016fast}, Gaussian mixture models \cite{b19,huh2017adaptive,huh2016learning}, and many other approaches\cite{burns2005toward}. Han et al. improved the efficiency and accuracy of prior approaches by decomposing the n-dimensional C-space into  subspaces and training a classifier like SVM or K-NN per each subspace to predict 1 if a configuration in the subspace is free 0 otherwise \cite{b3}. To determine whether a configuration $\vec{q}$ belongs to $C_\text{clsn}$ and $C_\text{free}$, the authors compute a logical 'and' amongst all classifiers \cite{b3}. Other recent learning based approaches in clude Fastron\cite{fastron_das2020learning} and its variations \cite{forward_kinematics_kernel_das2020forward,fastron_verghese2022configuration}, as well as DeepCollide, which makes use of implicit neural representations\cite{guo2023deepcollide}.

Our use of convolutional encoder-decoders is inspired by previous works that leveraged the technique to extract information from one image and represent that information via a new image. Molina et al. use a convolutional encoder-decoder to identify critical regions, regions of C-space that are unlikely to be sampled under SBMP techniques, but are critical for a number of motion planning problems (e.g. narrow corridors) \cite{b5}. For a SE(2) robot, the input to their model is a top-view image of a 3-D workspace and the output is the same top-view image augmented with green pixels to denote a critical region. Jardani et al. elevate the use of convolutional encoder-decoders beyond augmenting the input image with additional information. They use the technique to learn the function that maps a picture of a hydraulic transmissivity field to an image of its associated hydraulic tomography \cite{b6}. Much like Jardani et al., we utilize a convolutional encoder-decoder to learn the association between two visually different but conceptually related spaces, the robotic workspace and C-space.

Our work is also related to transfer learning and domain adaptation for convolutional neural networks (CNN). \cite{b7,b8} showed CNNs learn highly transferable features. We seek to evaluate how readily our convolutional encoder-decoder can adapt to workspaces that involve new types of transformations of obstacles.

\section{METHODOLOGY}

\subsection{Convolutional Encoder-Decoder Framework}
Inspired by \cite{b5,b6}, our method uses SegNet \cite{b10} to generate C-spaces from robotic workspaces. Initially designed for pixel-wise semantic scene segmentation, the SegNet model uses a convolutional encoder-decoder framework. The image of the robotic workspace is processed sequentially by a series of encoder blocks and corresponding decoder blocks. SegNet's encoder and decoder consist of 5 blocks each. The encoder and decoder is comprised of 13 convolutional layers for a total of 26 convolutional layers. The 13 convolutional layers in the encoder are topologically identical to the 13 convolutional layers of VGG16 \cite{b10,b11}. Our model extends the number of encoder and decoder blocks to 7 each. The encoder and decoder each have 19 convolutional layers for a total of 38 convolutional layers. We elected for 7 encoder and 7 decoder blocks after evaluating model performance on 5, 6, and 7 encoder and decoder blocks. The model with 7 encoder and 7 decoder blocks had the lowest validation loss. 

Each encoder and decoder block consists of either 2 or 3 pairs of a convolutional and batch normalization layer, followed by a ReLU nonlinearity. At the end of each encoder block is a max pool operation. At the start of each decoder block is a max unpooling operation. For a given decoder block, the max unpooling operation uses the pooling indices returned by the max pooling function at the end of the decoder's corresponding encoder block. All convolutional layers have a kernel size of 3 x 3 with a stride of 1. The max pool and unpool operations have a kernel size and stride of 2. Fig. 2 illustrates the network architecture.

\subsection{Loss Function}
The loss function is a pixel-wise regression loss between the predicted C-space and ground-truth C-space. Specifically, we optimize the weights, $\theta$, by alternating between an $L_2$ and $L_1$ pixel-wise regression loss between epochs: 

\begin{align}
    L_2(\theta) = \sum_{i=1}^{N}\sum_{p=1}^P (Y_{i,p} - \hat{Y}_{i,p})^2 \\
    L_1(\theta) = \sum_{i=1}^{N}\sum_{p=1}^P \left\lvert Y_{i,p} - \hat{Y}_{i,p} \right\rvert
\end{align}
where $N$ and $P$ denote the number of images per batch  and the number of pixels per image, respectively. Informed by loss functions standard in image restoration \cite{b14}, we chose to alternate between first minimizing the $L_2$ and the $L_1$ loss after also evaluating the model on the following loss functions: $L_1$, $L_2$, and binary cross entropy loss. Alternating between the $L_2$ and the $L_1$ loss functions had the lowest validation loss and lowest percentage of undetected collisions and undetected free space.

\section{EXPERIMENTS}

\begin{table*}[t]
\centering
\vspace*{5mm}
\caption{MAIN RESULTS}
\centering
 \begin{tabular}{| c c c c c c c |}
 \hline
 Dataset & Transformation &  Acc. (\%) & Prec. (\%) & Recall (\%) & F1 (\%) & Time $(\mu$s)  \\ 
 \hline
 3 circle  & Translation & 97.74 & 99.02 & 97.36 & 98.18 & 1.07\\
 \hline
 1-3 circle  & Translation, Removal & 98.15 & 99.15 & 98.76 & 99.14 & 1.07 \\
  \hline
 3 convex  & Translation & 95.83 & 98.24 & 94.50 & 96.33 & 1.10 \\
 \hline
  3 convex  
rotated & Translation, Rotation & 96.03 & 94.78 & 98.05 & 96.37 & 1.09 \\
 \hline
 \end{tabular}
\end{table*}
\begin{table*}
     \caption{CONFUSION MATRICES}
    \begin{subtable}[h]{0.5\textwidth}
    \centering
    \begin{tabular}{l|l|c|c|c}
    \multicolumn{2}{c}{}&\multicolumn{2}{c}{Predicted}&\\
    \cline{3-4}
    \multicolumn{2}{c|}{}&Collision&Free&\multicolumn{1}{c}{}\\
    \cline{2-4}
    \multirow{2}{*}{Actual}& Collision & $98.40 \%$ & $1.60 \%$ & \\
    \cline{2-4}
    & Free & $2.60\%$ & $97.40\%$ & \\
    \cline{2-4}
    \multicolumn{1}{c}{} & \multicolumn{1}{c}{} & \multicolumn{1}{c}{} & \multicolumn{1}{c}{} & \multicolumn{1}{c}{}\\
    \end{tabular}
    \caption{3 circle obstacles}
    \end{subtable}
   \begin{subtable}[h]{0.5\textwidth}
    \centering
    \begin{tabular}{l|l|c|c|c}
    \multicolumn{2}{c}{}&\multicolumn{2}{c}{Predicted}&\\
    \cline{3-4}
    \multicolumn{2}{c|}{}&Collision&Free&\multicolumn{1}{c}{}\\
    \cline{2-4}
    \multirow{2}{*}{Actual}& Collision & $97.91\%$ & $2.29\%$ & \\
    \cline{2-4}
    & Free & $0.82\%$ & $99.18\%$ & \\
    \cline{2-4}
    \multicolumn{1}{c}{} & \multicolumn{1}{c}{} & \multicolumn{1}{c}{} & \multicolumn{1}{c}{} & \multicolumn{1}{c}{}\\
    \end{tabular}
    \caption{1-3 circle obstacles}
    \end{subtable}
    
   \begin{subtable}[h]{0.5\textwidth}
    \centering
    \begin{tabular}{l|l|c|c|c}
    \multicolumn{2}{c}{}&\multicolumn{2}{c}{Predicted}&\\
    \cline{3-4}
    \multicolumn{2}{c|}{}&Collision&Free&\multicolumn{1}{c}{}\\
    \cline{2-4}
    \multirow{2}{*}{Actual}& Collision & $97.73\%$ & $02.27\%$ & \\
    \cline{2-4}
    & Free & $5.50\%$ & $94.50\%$ & \\
    \cline{2-4}
    \multicolumn{1}{c}{} & \multicolumn{1}{c}{} & \multicolumn{1}{c}{} & \multicolumn{1}{c}{} & \multicolumn{1}{c}{}\\
\end{tabular}
        \caption{3 convex obstacles}
    \end{subtable}
       \begin{subtable}[h]{0.5\textwidth}
    \centering
\begin{tabular}{l|l|c|c|c}
\multicolumn{2}{c}{}&\multicolumn{2}{c}{Predicted}&\\
\cline{3-4}
\multicolumn{2}{c|}{}&Collision&Free&\multicolumn{1}{c}{}\\
\cline{2-4}
\multirow{2}{*}{Actual}& Collision & $97.57\%$ & $2.43\%$ & \\
\cline{2-4}
& Free & $5.16\%$ & $94.84\%$ & \\
\cline{2-4}
\multicolumn{1}{c}{} & \multicolumn{1}{c}{} & \multicolumn{1}{c}{} & \multicolumn{1}{c}{} & \multicolumn{1}{c}{}\\
\end{tabular}
        \caption{3 convex obstacles rotated}
    \end{subtable}     
\end{table*}
\subsection{Dataset}
\begin{figure}[h]
    \centering
    \begin{minipage}[b]{0.45\textwidth}
        \centering
        \includegraphics[scale=0.35]{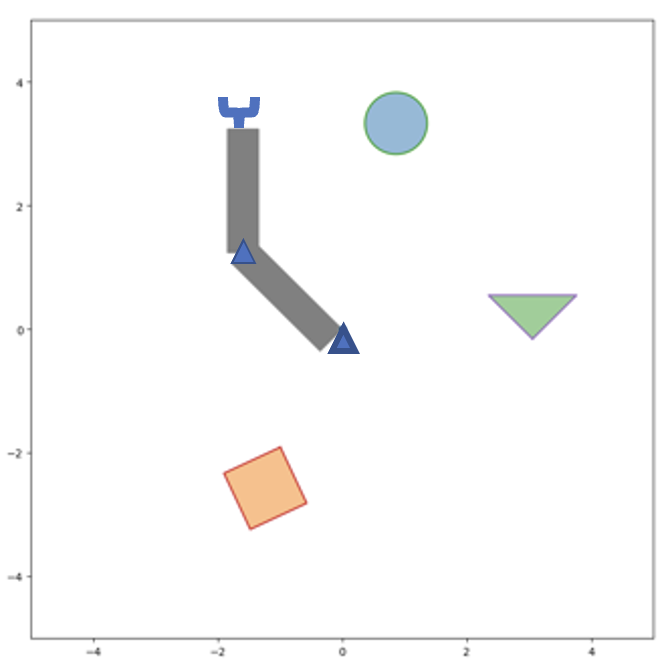}
    \end{minipage}%
    \hfill
    \begin{minipage}[b]{0.5\textwidth}
        \captionof{figure}{Sample workspace from 3 convex obstacles rotated dataset. The gray boxes represent the rigid dual-arm robot. The arms can rotate freely, but the robot cannot perform translational movement. The circle, square, and triangle are obstacles in the workspace.}
        \label{fig:workspace}
    \end{minipage}
\end{figure}

We curate four datasets to the study convolutional encoder-decoder techniques for C-space prediction. Each dataset consists of 10k 2-D robotic workspace, C-space image pairs for a dual-arm robot with 2 joints of freedom, $Q_1$ and $Q_2$, illustrated in  Fig. \ref{fig:workspace}. The rigid body robot cannot perform translational motion, but each arm is allowed to freely rotate $360^{\circ}$. Each image of the robotic workspace consists of only the obstacles. Since each joint's range is constant, we remove the axes describing each joint. We include collisions arising from the robotic arms' mutual interference as the arms are expected to accomplish their task without colliding with one another \cite{b15,b16}. 

\begin{figure}[h]
\vspace*{2mm}
     \centering
     \begin{subfigure}[]{.225\textwidth}
         \centering
         \includegraphics[width=\textwidth]{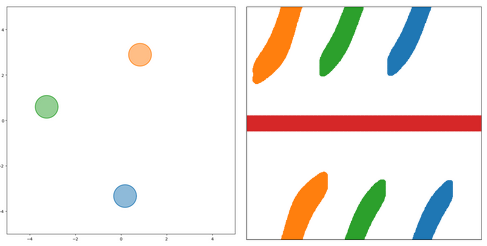}
         \caption{3 circle obstacles}
         \label{fig:3_circles}
     \end{subfigure}
           \hfill
         \begin{subfigure}[]{.225\textwidth}
         \centering
\includegraphics[width=\textwidth]{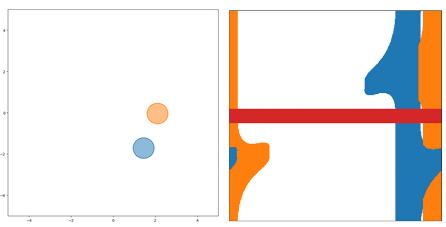}
         \caption{1-3 circle obstacles}
         \label{fig:1_3_circles}
     \end{subfigure}
           \hfill
      \begin{subfigure}[]{.225\textwidth}
         \centering
         \includegraphics[width=\textwidth]{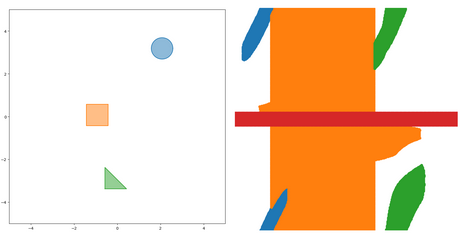}
         \caption{3 convex obstacles}
         \label{fig:3_convex_shapes}
     \end{subfigure}
      \hfill
     \begin{subfigure}[]{.225\textwidth}
         \centering
         \includegraphics[width=\textwidth]{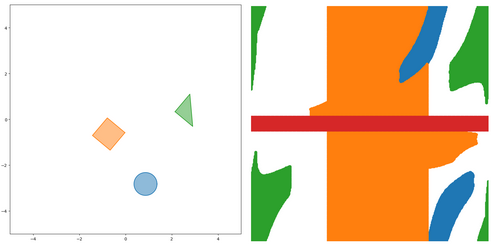}
         \caption{3 convex obstacles rotated}
         \label{fig:3_convex_rotate}
     \end{subfigure}
        \caption{Sample workspace and C-space image pairs from their respective datasets. Color is used to associate the obstacles' contributions to the C-space. Red symbolizes configurations that result in self-collision.}
\end{figure}

 Each dataset tests a different setting for C-space prediction. The first dataset, 3 circle obstacles, consists of 3 obstacles of the same shape and size randomly translated throughout the workspace. The second dataset, 1-3 circle obstacles, builds upon the first by randomly removing 0, 1, or 2 obstacles from the workspace. The third, 3 convex obstacles, includes 3 different convex shapes (square, circle, triangle) randomly placed throughout the workspace. The fourth, 3 convex obstacles rotated, builds upon the third by randomly rotating the different convex shapes in the workspace. Fig. 4 displays sample workspace, C-space pairs for all four datasets.

\subsection{Settings}
We train our convolutional encoder-decoder for 60 epochs or until validation loss plateaus using AdaDelta \cite{b13} with a learning rate of 0.01. We did not use weight decay regularization or gradient clipping. The batch normalization layers in the model serve as a form of regularization \cite{b12}. At the end of every 25 epochs, we decay the learning rate by 0.75. We used a 0.7, 0.15, and 0.15 split for train, test, and validation sets. We use a batch size of 5 image pairs for training. We converted each robotic workspace and ground-truth robotic C-space image to black and white and scaled each image to a 512 x 512 image. A black pixel and white pixel denote a collision and no-collision, respectively. As the model is trained to minimize the $L_2$ and $L_1$ regression losses, during inference, we used a threshold, $\eta$, to convert the pixel-value predictions to binary values. We determined $\eta$ by testing different threshold values and selecting the value that performed the best on the validation set. Pixel values less than $\eta$ were set to 0, denoting a collision, and greater than or equal to $\eta$ were set to 1, indicating a free configuration. We trained the model on an NVIDIA 2080-TI and it converged to its lowest loss after 7 hours of training. The model has 33.87 million trainable parameters compared to SegNet's 29.4 million trainable parameters \cite{b10}.

\begin{figure*}[]
\vspace*{5mm}
     \centering
     \begin{subfigure}[]{.8\textwidth}
         \centering  \includegraphics[width=\textwidth]{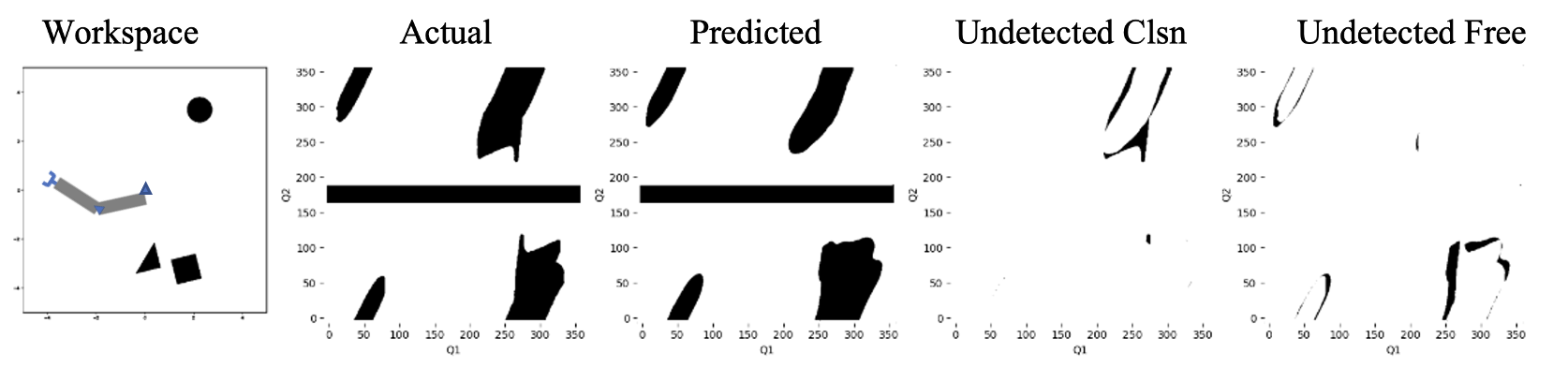}
         \caption{1750 image pairs }
     \end{subfigure}
     \hfill
     \begin{subfigure}[]{.8\textwidth}
         \centering
         \includegraphics[width=\textwidth]{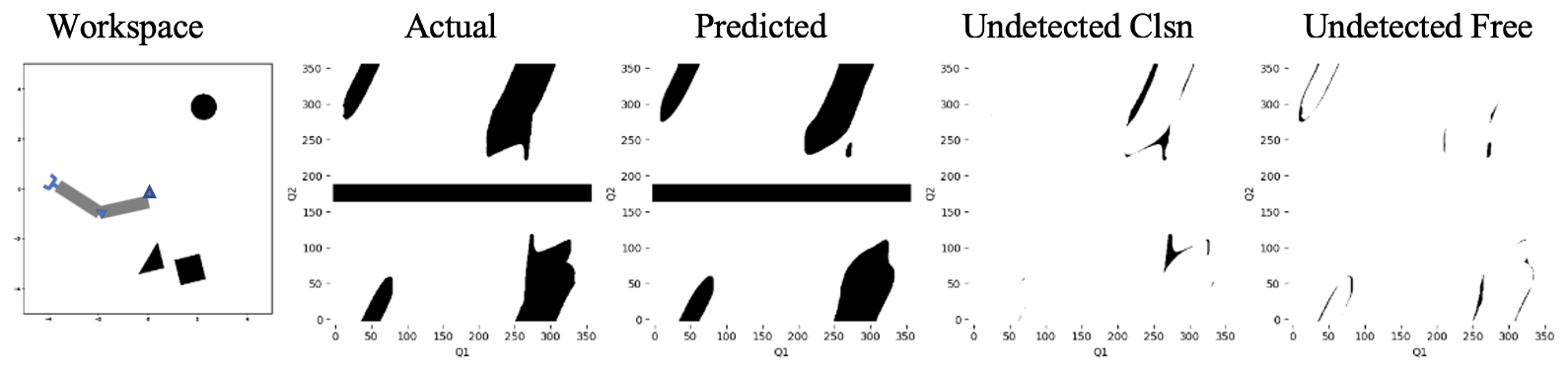}
         \caption{3500 image pairs}
     \end{subfigure}
      \hfill
     \begin{subfigure}[]{.8\textwidth}
         \centering
         \includegraphics[width=\textwidth]{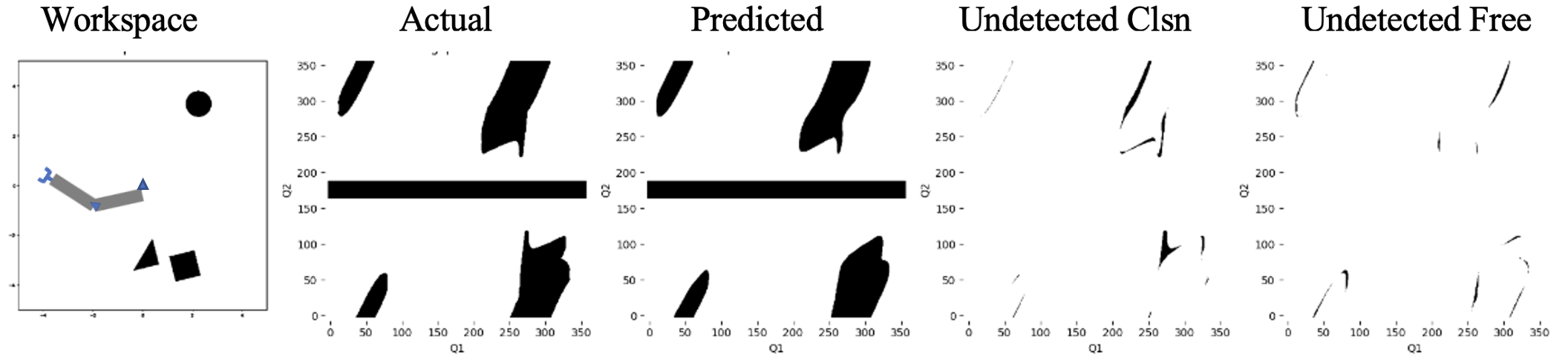}
         \caption{5250 image pairs}
     \end{subfigure}
      \hfill
     \begin{subfigure}[]{.8\textwidth}
         \centering
         \includegraphics[width=\textwidth]{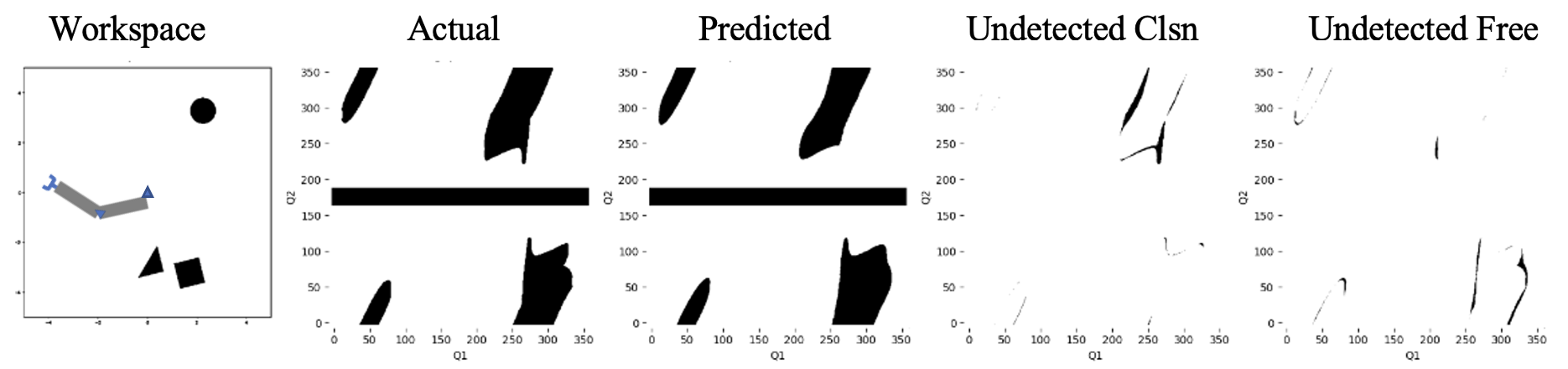}
         \caption{7000 image pairs}
     \end{subfigure}
        \caption{Constructed C-spaces as a function of the number of workspaces seen. In order from left to right: workspace, actual, predicted, undetected collisions, and undetected free space. Black pixels indicate undetected collisions and undetected free space in their respective plots. By increasing the number of image pairs, the model better detects the boundary between $\text{C}_{\text{free}}$ and $\text{C}_{\text{clsn}}$.}
\end{figure*}

\section{RESULTS \& DISCUSSION}
\subsection{Main Results}

Table I summarizes our main results on the respective dataset test sets. We report our model's pixel-wise accuracy, precision, recall, and F1 scores of the prediction of $C_\text{clsn}$ and $C_\text{free}$. We also report the time to construct the C-space per configuration in microseconds ($\mu s$). We first trained two independent copies of the model on the 3 circle obstacle and 3 convex obstacle datasets to obtain the reported metrics. \cite{b15} demonstrated CNN features are highly transferable across domains. While more complex methods for CNN domain adaption exist, to adapt the instance of the model trained on the 3 circle obstacle dataset to the 1-3 circle obstacle dataset, we fine-tuned the model for 20 epochs at a .001 learning rate. We performed the same procedure for adapting the model trained on the 3 convex shape dataset to the 3 convex shape dataset with rotations. 

As reported, the model achieves consistently high metrics with an average F1-score of 97.50\% over all datasets. For the domain-adapted datasets, 1-3 circle obstacles and 3 convex obstacles rotated, the model obtains metrics comparable to those datasets the model was originally trained on. Prior SBMP techniques report the collision status of a single configuration of a 2-DOF robot in 9 $\mu $s \cite{b3}. Our method constructs the complete C-space for a 2-DOF robot in 1.08 $\mu$s on average. Fig. 6 displays a sample workspace, ground-truth C-space, and predicted C-space triple for each dataset.

Quantifying the number of undetected collisions is paramount to the safety implications of our model. Therefore, we report the confusion matrices for the test sets of each dataset in Table II. Our model limits the number of undetected collisions to 2.15\% on average. Increasing the threshold, $\eta$, at which a configuration is classified as free decreases the number of undetected collisions at the cost of increasing the amount of undetected free space. 

\begin{figure}[ht]
    \centering
    \begin{subfigure}[b]{0.48\textwidth}
        \centering
        \includegraphics[width=\textwidth]{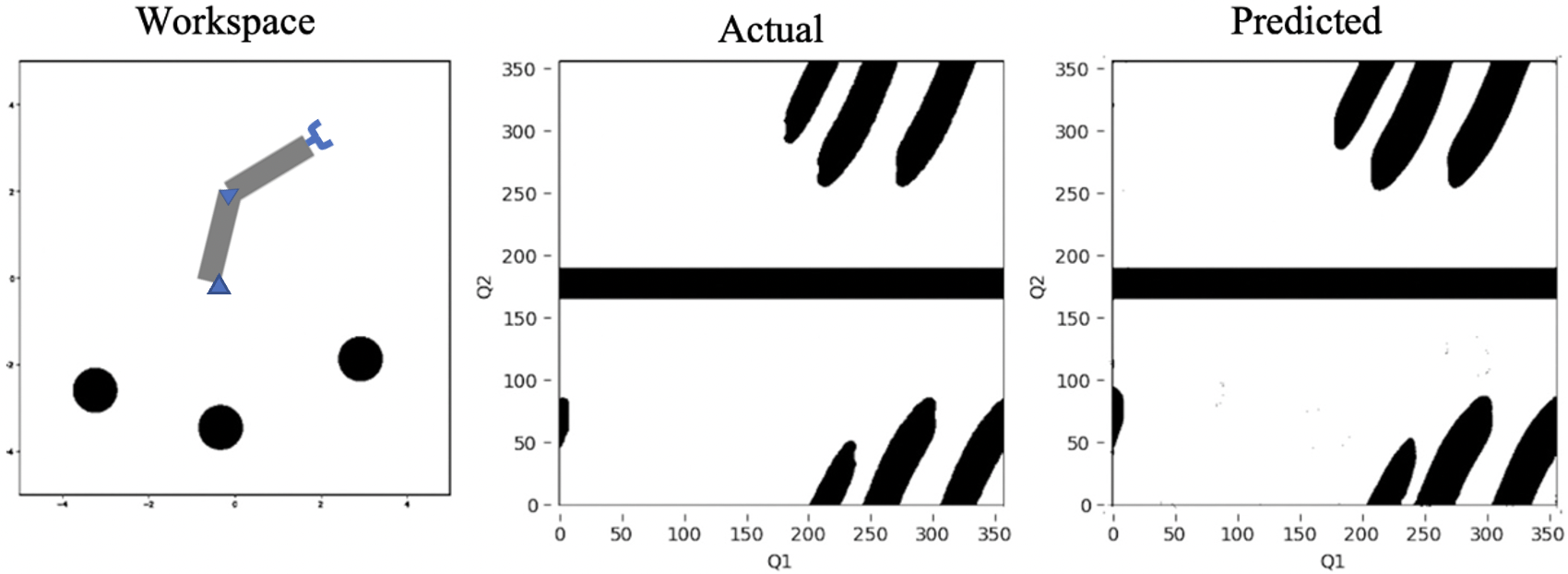}
        \caption{3 circle obstacles}
    \end{subfigure}
    \hfill
    \begin{subfigure}[b]{0.48\textwidth}
        \centering
        \includegraphics[width=\textwidth]{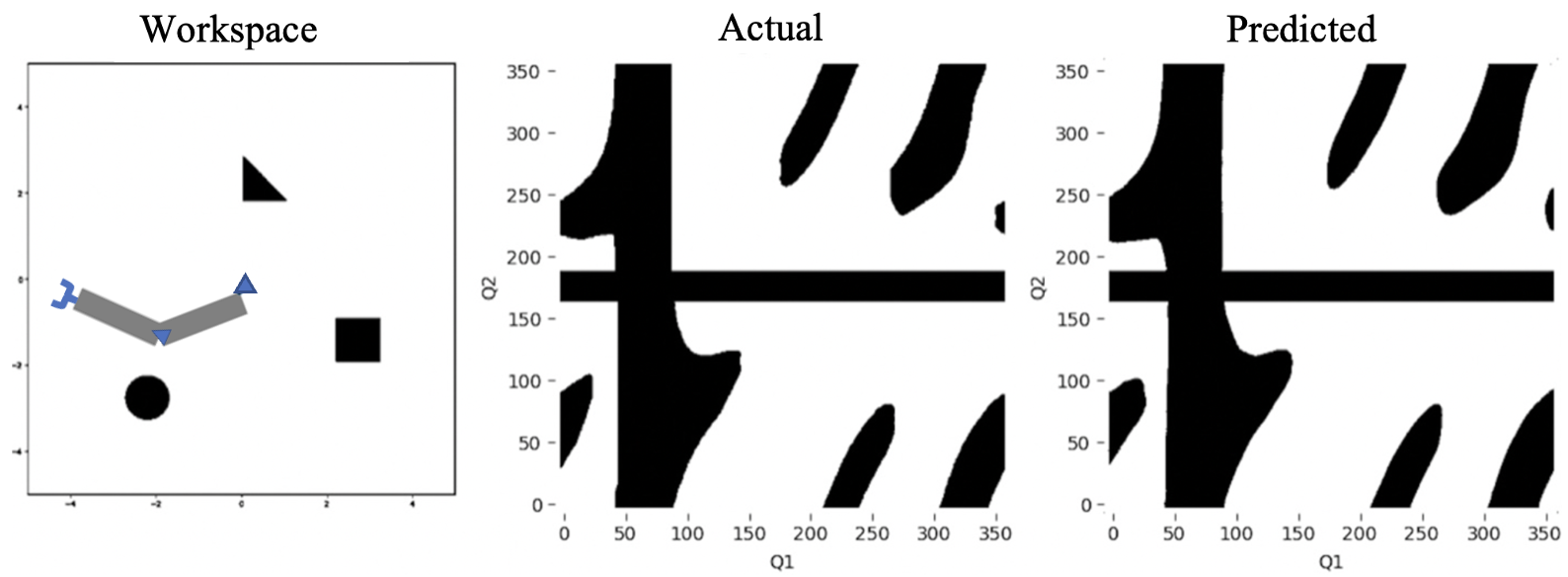}
        \caption{3 convex obstacles}
    \end{subfigure}
    \vskip\baselineskip
    \begin{subfigure}[b]{0.48\textwidth}
        \centering
        \includegraphics[width=\textwidth]{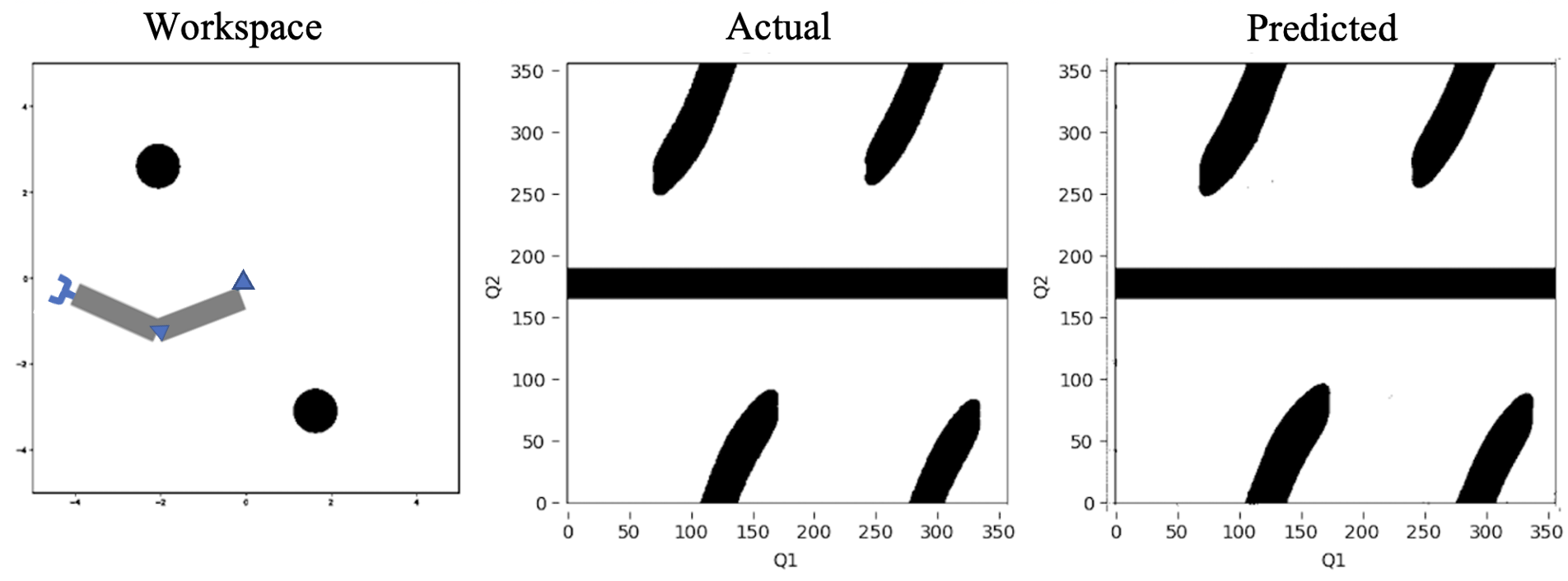}
        \caption{1-3 circle obstacles}
    \end{subfigure}
    \hfill
    \begin{subfigure}[b]{0.48\textwidth}
        \centering
        \includegraphics[width=\textwidth]{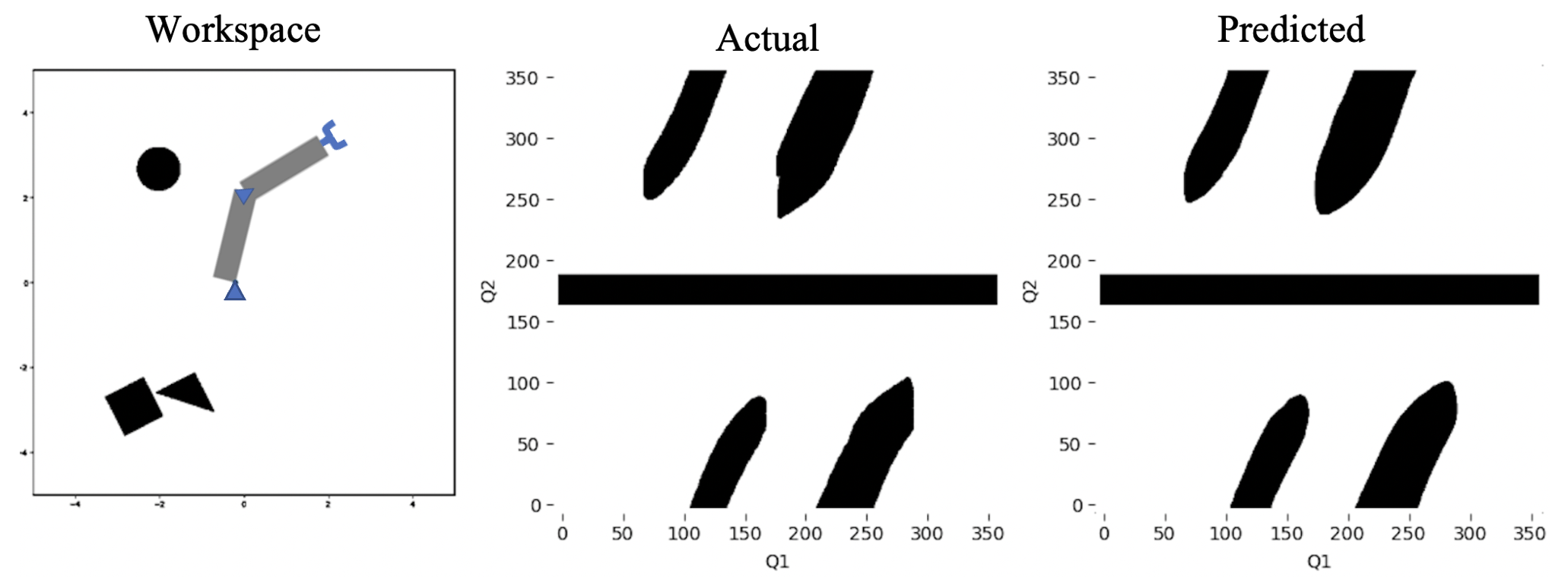}
        \caption{3 convex obstacles rotated}
    \end{subfigure}
    \caption{Sample C-space constructions for each dataset.}
    \label{fig:robotic-workspace-with-robot}
\end{figure}

\subsection{Effect of Training Data Size}

We examine the impact of the number of workspaces seen on the ability of our model to construct highly accurate C-spaces from workspaces. For this analysis, we provide \\ $n=1750, 3500, 5000, \text{and the full } 7000$ robotic workspace, C-space image pairs during training on the 3 convex obstacles rotated dataset. We evaluate using the standard $1500$ pairs of robotic workspaces and C-spaces for both the validation and test sets. During training, we stop the model early if the model's validation loss has stayed the same in 4 epochs to avoid over-fitting in the cases of limited data.

Table \ref{Robotic-workplaces-seen} shows the relationship between the number of robotic workplaces seen during training and the performance on key metrics, F1-score, and percentage of undetected collisions and undetected free space. Fig. 5 displays constructed C-spaces for the different numbers of training samples provided.

\begin{table}[h!]
\centering
 \caption{EFFECT OF NUMBER OF WORKSPACE C-SPACE PAIRS SEEN}
 \begin{tabular}{|c c c c|} 
 \hline
Samples & F1 (\%) & Missed Clsn (\%) & Missed Free (\%) \\ 
 \hline
1750 & 94.53 & 8.95 & 2.55 \\
 \hline
3500 & 95.64 & 4.81 & 3.70 \\
  \hline
  5250 & 95.69 & 6.58 & 2.35\\
 \hline
  7000 & 96.37 & 2.29 & 5.16\\
 \hline
 \end{tabular} \\
 \label{Robotic-workplaces-seen}
\end{table}

While training on the complete set of $7000$ image pairs achieved the best results, the model obtained a $94.33 \%$ F1-score with less than $9\%$ undetected collisions using only 1750 robotic workspace, C-space image pairs. The low percentage of undetected free space and a comparatively higher percentage of undetected collisions suggests the model quickly detects free space at the cost of detecting collisions. As illustrated in the undetected collisions and undetected free space plots of Fig. 5, by providing additional workspace, C-space pairs, the model can further refine the boundary between $\text{C}_{\text{free}}$ and $\text{C}_{\text{clsn}}$, decreasing the number of undetected collisions and the number of undetected free space. 

\subsection{Transferability of Features}
We are interested in how our method adapts to new transformations of obstacles in robotic workspaces. While our primary results demonstrated that our technique can achieve strong results through fine-tuning, we are interested in exploring how the method performs under a zero-shot setting.

\begin{figure}[]
\vspace*{2mm}
     \centering
     \begin{subfigure}[b]{0.45\textwidth}
         \centering
         \includegraphics[width=\textwidth]{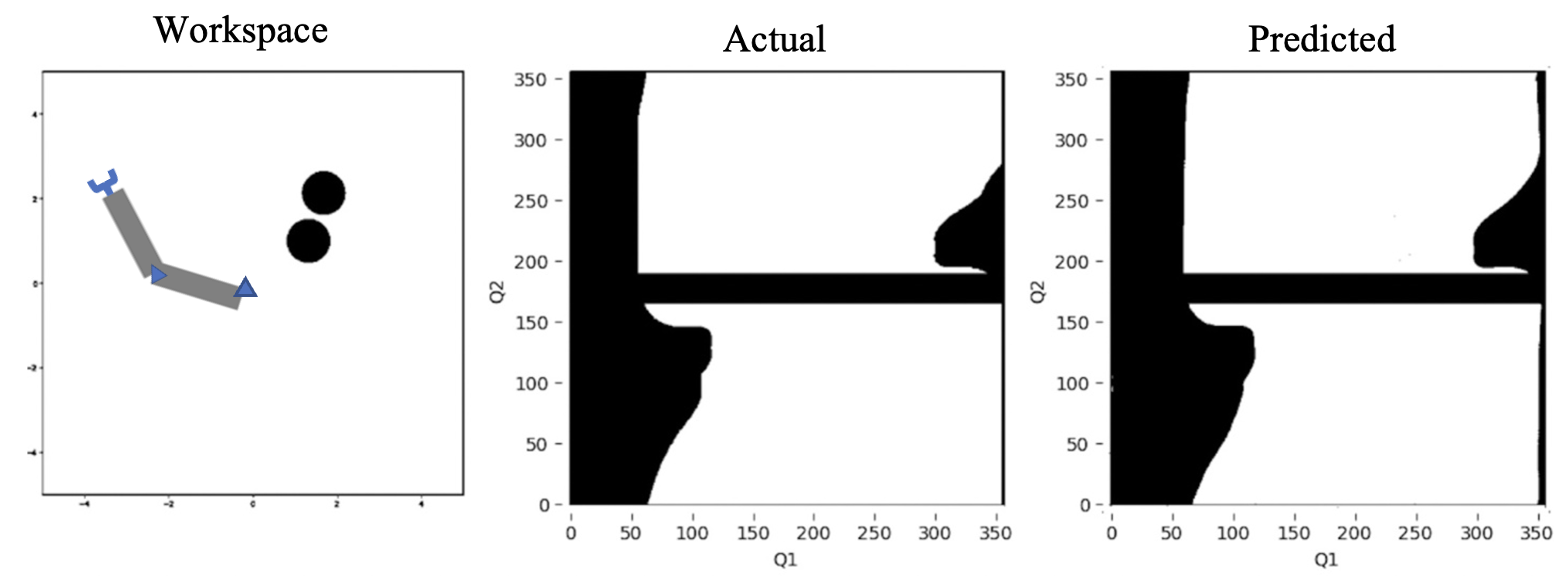}
         \caption{1-3 circle obstacles}
     \end{subfigure}
    \begin{subfigure}[b]{0.45 \textwidth}
         \centering
         \includegraphics[width=\textwidth]{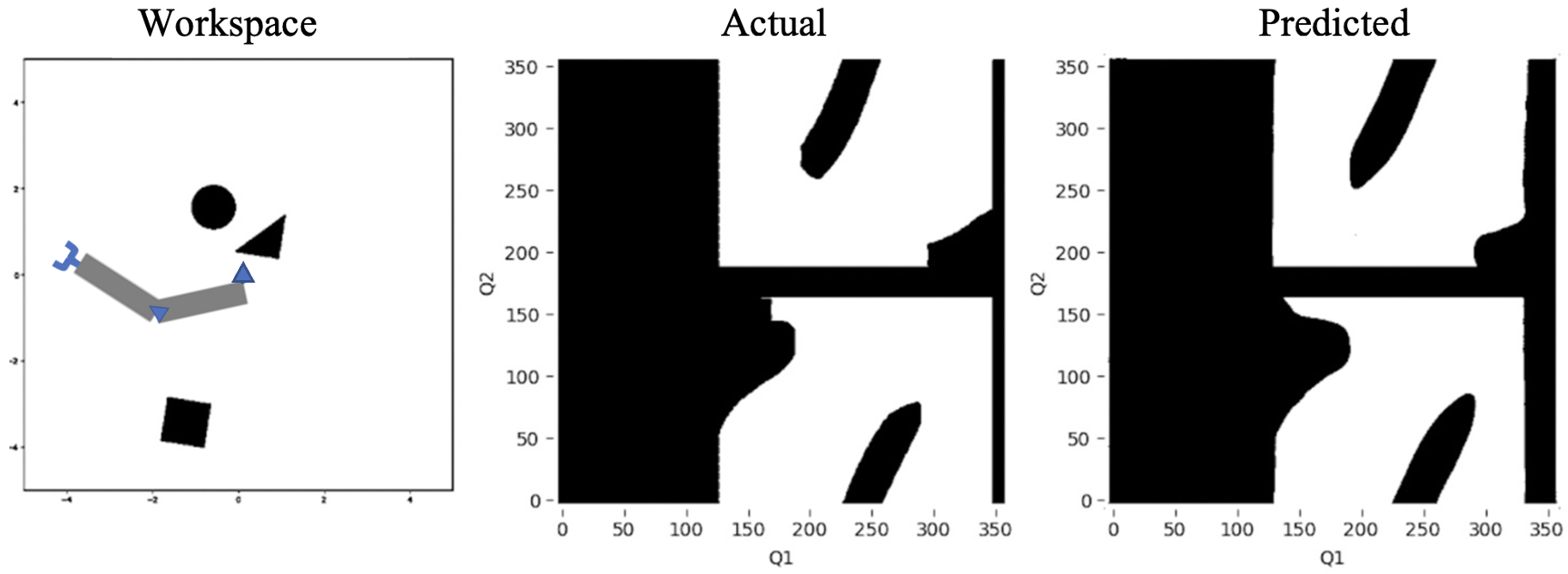}
         \caption{3 convex obstacles rotated}
     \end{subfigure}
     \hfill
        \caption{Zero-shot images.}
        \label{fig:zero-shot}
\end{figure}

We evaluate the model trained on the 3 circle obstacles dataset on the 1-3 circle obstacles dataset in a zero-shot setting. We perform the same experiment but with the model trained on the 3 convex obstacles on the 3 convex obstacles rotated dataset. We provide the complete results in Table IV. Our convolutional encoder-decoder still achieves an average 97.01\% F1-score without training on the removal and rotations transformations from the new datasets. Therefore, our method learns highly transferable features that can be applied to workspaces that involve new transformations. Sample C-space predictions are provided in Fig. \ref{fig:zero-shot}.

\begin{table}[h!]
  \caption{ZERO-SHOT PERFORMANCE}
  \centering
  \small
  \begin{tabular}{|c c c c|} 
    \hline
    Transformation & F1 (\%) & Missed Clsn (\%) & Missed Free (\%) \\ 
    \hline
    Removal   & 97.95 & 0.78 & 3.61 \\
    \hline
    Rotation  & 96.07 & 3.69 & 5.03 \\
    \hline
  \end{tabular}
\end{table}

\section{CONCLUSION}
In this work, we proposed a novel method for direct  C-space construction from robotic workspaces using convolutional encoder-decoders. Unlike previous statistical based motion planning (SMBP) approaches that train on sampled configuration and collision status pairs, our technique trains on workspace and C-space pairs. Doing so allows our model to predict highly accurate approximations to C-space for dynamic workspaces directly. Our model is able to adapt to new obstacle transformations through little to no fine-tuning. Future works may consider applying this technique to constructing C-spaces for 3-D workspaces or consider incorporating convolutional encoders into SMBP procedures.

\bibliography{example_paper}
\bibliographystyle{icml2025}


\end{document}